\documentclass[10pt,twocolumn,letterpaper]{article}

\usepackage{iccv}
\usepackage{times}
\usepackage{epsfig}
\usepackage{graphicx}
\usepackage{amsmath}
\usepackage{amssymb}
\usepackage{multirow}
\usepackage{pifont}

\usepackage[breaklinks=true,bookmarks=false]{hyperref}

\iccvfinalcopy 

\def\iccvPaperID{****} 

\ificcvfinal\fi

\def\blu#1{\textbf{\color{blue} #1}} 
\def\red#1{\textbf{\color{red}\underline{#1}}} 
\begin{document}

\title{Light Field Saliency Detection with Dual Local Graph Learning and Reciprocative Guidance}

\author{
Nian Liu$^{1}$\footnotemark[1]\quad
Wangbo Zhao$^{2}$\footnotemark[1]\quad
Dingwen Zhang$^{2}$\quad
Junwei Han$^{2}$\footnotemark[2]\quad
Ling Shao$^{1}$\quad\\
$^1$ Inception Institute of Artificial Intelligence\quad $^2$ Northwestern Polytechnical University \quad\\
    {\tt\small
    \{liunian228, wangbo.zhao96, zhangdingwen2006yyy, junweihan2010\}@gmail.com,}\\
    {\tt\small ling.shao@ieee.org
    }
}
\maketitle
\footnotetext[1]{Equal contribution.}
\footnotetext[2]{Corresponding author.}

\maketitle
\ificcvfinal\thispagestyle{empty}\fi

\begin{abstract}
The application of light field data in salient object detection is becoming increasingly popular recently. The difficulty lies in how to effectively fuse the features within the focal stack and how to cooperate them with the feature of the all-focus image. Previous methods usually fuse focal stack features via convolution or ConvLSTM, which are both less effective and ill-posed. In this paper, we model the information fusion within focal stack via graph networks. They introduce powerful context propagation from neighbouring nodes and also avoid ill-posed implementations. On the one hand, we construct local graph connections thus avoiding prohibitive computational costs of traditional graph networks. On the other hand, instead of processing the two kinds of data separately, we build a novel dual graph model to guide the focal stack fusion process using all-focus patterns. To handle the second difficulty, previous methods usually implement one-shot fusion for focal stack and all-focus features, hence lacking a thorough exploration of their supplements. We introduce a reciprocative guidance scheme and enable mutual guidance between these two kinds of information at multiple steps. As such, both kinds of features can be enhanced iteratively, finally benefiting the saliency prediction. Extensive experimental results show that the proposed models are all beneficial and we achieve significantly better results than state-of-the-art methods.
\end{abstract}

\section{Introduction}
Salient object detection (SOD) methods
can be categorized into RGB based ones, RGB-D based ones, and the recently proposed light field based ones. By only based on static images, although RGB SOD methods  \cite{hou2017deeply, liu2020picanet, liu2019simple, zhao2020suppress} have achieved excellent performance on many benchmark datasets, they still can not handle challenging and complex scenes. This is because the appearance saliency cues conveyed in RGB images are heavily constrained, especially when the foreground and background appearance are complex or similar. To solve this problem, depth information is introduced to provide supplementary cues in RGB-D SOD methods  \cite{chen2018progressively, zhao2019contrast, piao2019depth, liu2020learning}. However, it is not easy to obtain high-quality depth maps and many current RGB-D SOD benchmark datasets only have noisy depth maps.
On the contrary, light field data 
Hence, the light field SOD problem has much potential to explore. 

\begin{figure}[!t]
	\graphicspath{{figure/}}
	\centering
	\includegraphics[width=1\linewidth]{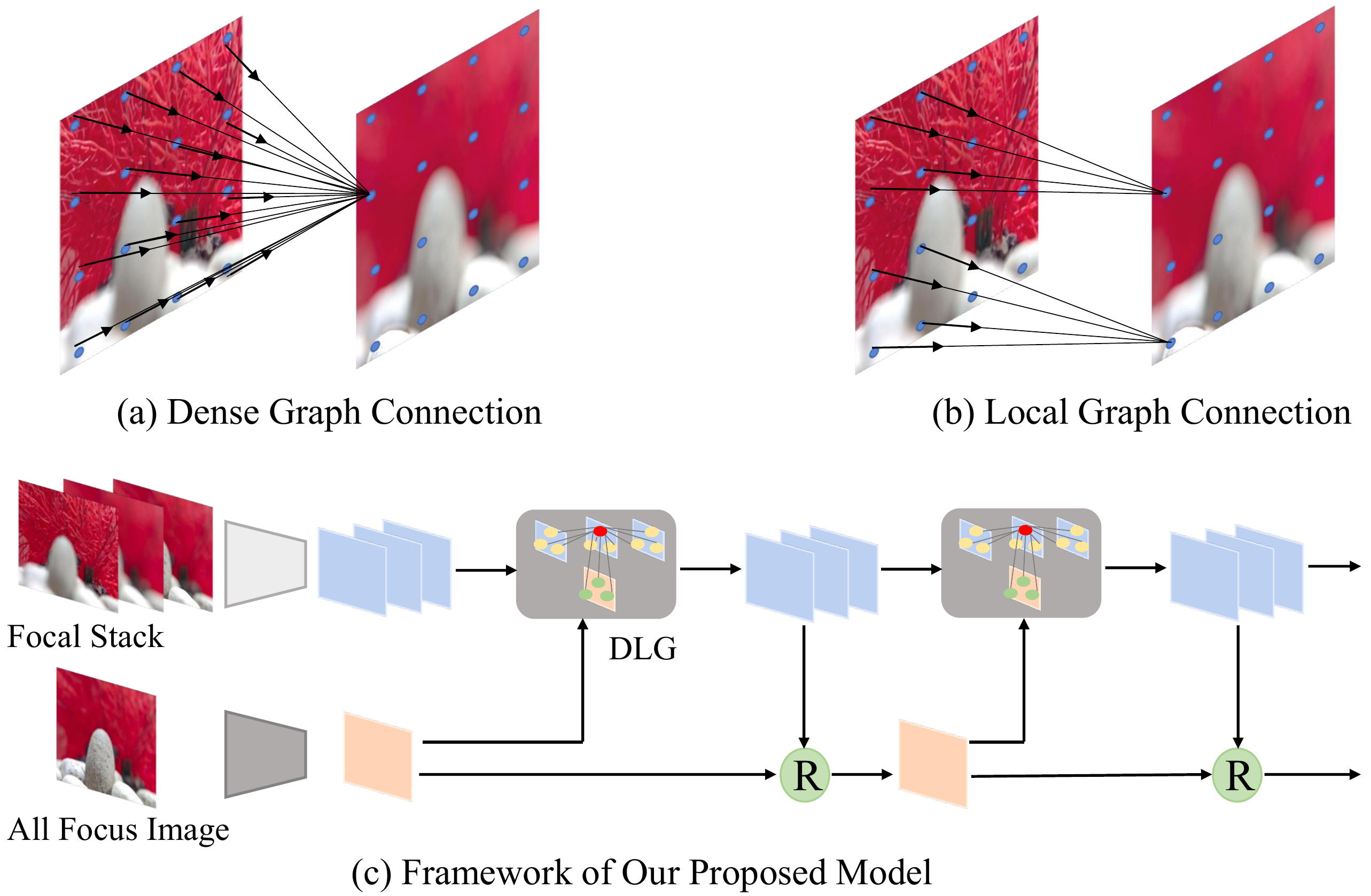}
	\caption{(a) and (b) show the comparison of traditional dense graph models and our proposed local graph model. (c) illustrates the framework of our model. \begin{normalsize}{\textcircled{\footnotesize{R}}}\end{normalsize} means the reciprocative unit.}
	\label{fig:figure1}
	\vspace{-0.5cm}
\end{figure}

Besides the focal stack images, light field data also have an all-focus image that provides the context information. Thus, light field SOD has two key points, \ie, \emph{how to effectively fuse multiple focal stack features and how to cooperate the focal stack cues with the all-focus information}. A straightforward way to solve the first problem is to concatenate focal stack features and use a convolution layer for fusion. Such a simple way can not sufficiently explore the complex interaction within different focal slices, hence may limit the model performance.
It is also an ill-posed solution since convolution requires a fixed input number, hence many methods have to randomly pad the input images when they are less than the pre-defined number.
Adopting ConvLSTM  \cite{shi2015convolutional} is another popular solution,
where the focal stack images are processed one by one in a pre-defined sequential order using the memory mechanism. This also involves an ill-posed problem setting since there is no meaningful order among focal stack images. Furthermore, the usage of the sequential order may cause ConvLSTM to
ignore the information of the focal slices that are input earlier. 
As for the second problem, most previous works \cite{wang2019dlsd} simply concatenate or sum the focal stack feature with the all-focus feature and then adopt convolutional fusion only once. 
Such a straightforward fusion method heavily limits the exploration of complex supplementary relations between these two kinds of information.

To solve the first problem, adopting the powerful graph neural networks (GNNs)  \cite{gori2005new,scarselli2008graph} is a possible way. GNNs aggregate the contextual information from neighbouring nodes and propagate it to the target node, 
thus can achieve effective feature fusion. 
At the same time, it avoids the ill-posed implementation problem since the graph connection can be built flexibly and does not depend on a sequential order.
A straightforward way is to view each pixel location in the feature maps of the focal stack as a node and construct dense edge connections among all locations, as shown in Figure~\ref{fig:figure1}(a).
However, this is impractical since light field SOD requires large feature maps for focal slices to obtain fine-grained segmentation.
Hence, building a densely connected graph involves prohibitive computational costs.


To this end, we propose to build local graphs to efficiently aggregate contexts in different focal slices. 
We treat each image pixel location in the focal stack as nodes and build the graph only within local neighbouring nodes, as shown in Figure~\ref{fig:figure1}(b). As such, the context propagation within focal slices can be efficiently performed with dramatically reduced edge connections. 
One can further introduce multiscale local neighbours, hence incorporating larger context information with acceptable computational costs. 
Besides building a graph within the focal stack, we also build a focal-all graph to introduce external guidance from the all-focus feature for the fusion of focal features, thus resulting in a novel dual local graph (DLG) network.

To tackle the second key point in light field SOD, we propose a novel reciprocative guidance architecture, as shown in Figure~\ref{fig:figure1}(c). It introduces multi-step guidance between the all-focus image feature and the focal stack features. 
In each step, the former is first used to guide the fusion of the latter, and then the fused feature is used to update the former.
We perform such a process in a reciprocative fashion, where mutual guidance can be conducted recurrently. Finally, the two kinds of features can be improved with more discriminability, benefiting the final SOD decision.

Our main contributions can be summarised as:

\vspace{-3mm}
\begin{itemize}
    \item We propose a new GNN model named dual local graph to enable effective context propagation in focal stack features under the guidance of the all-focus feature and also avoid high computational costs.
\vspace{-1mm}
    \item We propose a novel reciprocative guidance scheme to make the focal stack and the all-focus features guide and promote each other at multiple steps, thus gradually improving the saliency detection performance. 
\vspace{-1mm}  
    \item Extensive experiments illustrate the effectiveness of our method. It surpasses other light field methods by a large margin. Moreover, with much less training data, our method also shows competitive or better performance compared with RGB-D or RGB based SOD models.

\end{itemize}

\section{Related Work}

\subsection{Light Field SOD}
Although the usage of CNNs has improved RGB SOD and RGB-D SOD by a large margin \cite{wang2021salient,zhou2021rgb}, there are still lots of challenges in the SOD task, especially when the visual scenes are complex. 
Hence, several works have tried to leverage the focal cues in light field data to perform SOD. \cite{li2014saliency} was the first work to explore SOD with light field data, and constructed the first benchmark dataset. After that, the background prior \cite{zhang2015saliency}, weighted sparse coding \cite{li2015weighted}, and light field flow \cite{zhang2017saliency} are widely used for this new task.
More details about traditional methods can be found in \cite{Fu2020lightSOD}.

When it comes to the deep learning era, several deep-learning methods have promoted the light field SOD performance signiﬁcantly.
Zhang \etal \cite{zhang2019memory} inputted the feature maps of focal slices and the all-focus image into a ConvLSTM  \cite{shi2015convolutional} to fuse them one by one. This scheme has the ill-posed implementation problem. On the other hand, their method only fuses focal stack and all-focus features once.
Wang \etal \cite{wang2019dlsd} and Piao \etal \cite{piao2020exploit} both fused the features from different focal slices using varying attention weights, which are inferred at multiple time steps in a ConvLSTM. As such, they performed feature fusion within focal slices several times.
However, \cite{wang2019dlsd} conducted focal stack and all-focus feature fusion only once, while \cite{piao2020exploit} did not perform such a fusion. They adopted knowledge distillation  \cite{hinton2015distilling} to improve the representation ability of the all-focus branch. 


Different from previous works, our proposed DLG network enables efficient context fusion among all focal slice images. We also introduce the guidance from the all-focus feature into the focal stack fusion process, and our proposed reciprocative architecture introduces mutual guidance between the two kinds of features multiple times.
These two points have never been explored by previous works.

\begin{figure*}[!t]
	\graphicspath{{figure/}}
	\centering
	\includegraphics[width=1\linewidth]{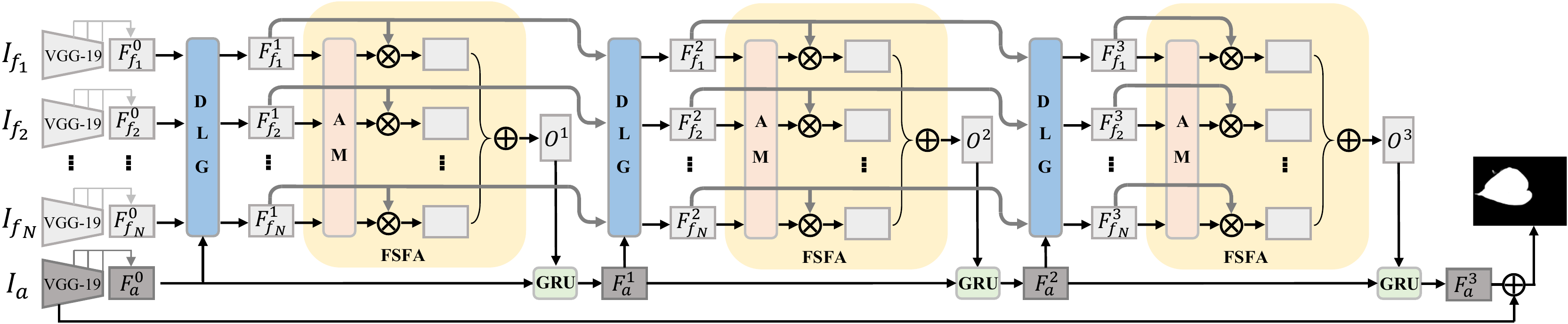}
	\caption{Overview of our proposed model. DLG: the proposed dual local graph; FSFA: focal stack feature aggregation; AM: used to generate the attention matrix $A$ in \eqref{eq:fsfa}; $\otimes$: element-wise multiplicationl; $\oplus$: element-wise addition; GRU: ConvGRU \cite{cho2014learning}. Due to space limitation, we only illustrate the reciprocative guidance process at three time steps.}
	\label{network_overview}
	\vspace{-0.4cm}
\end{figure*}

\subsection{Grapth Neural Network}
Graph neural networks (GNNs) were proposed by  \cite{gori2005new} and developed by  \cite{scarselli2008graph} to model data structures in graph domains. Since GNNs can model the relationships among nodes, they have been applied in many fields, such as molecular biology \cite{gilmer2017neural}, natural language processing \cite{beck2018graph}, knowledge graph \cite{hamaguchi2017knowledge}, and disease classification \cite{rhee2017hybrid}.

Recently, GNNs have also been widely explored in the field of computer vision. Wang \etal \cite{wang2018videos} adopted a graph convolutional network to build the spatial-temporal relationships for action recognition. For dense prediction tasks, Luo \etal \cite{luo2020cascade} used GNNs to construct graphs among feature maps and learn cross-modality and cross-scale reasoning simultaneously for RGB-D SOD. In \cite{wang2019zero}, Wang \etal proposed an attentive GNN to learn the semantic and appearance relationships among several video frames for video object segmentation. Zhang \etal \cite{zhang2020adaptive} adopted a graph convolutional network to jointly implement both intra-saliency detection and inter-image correspondence for co-saliency detection. Both the latter two works constructed densely-connected pixel-pixel graphs, which are computationally expensive and lack scalability, especially for the light field data that can have more than 10 focal stack images. On the contrary, we propose a novel graph architecture with local pixel-pixel connections for light field SOD. We also introduce dilated neighbouring connections to incorporate large contexts with computational efficiency. Furthermore, we build two graphs to simultaneously propagate context interaction among focal stack images and incorporate the guidance from the all-focus image.

\subsection{Reciprocative Models}
Reciprocative or recurrent models, including RNN, LSTM \cite{hochreiter1997lstm}, and GRU \cite{cho2014gru}, process temporal sequences using internal state or memory, and progressively update their states. There are also many other saliency detection works or other related tasks using reciprocative models. AGNN \cite{wang2019zero} and CAS-GNN \cite{luo2020cascade} used reciprocative models as the node updating function in GNNs to update graph node embeddings. DMRA \cite{piao2019depth}, DLSD \cite{wang2019dlsd}, ERNet \cite{piao2020exploit}, and MoLF \cite{zhang2019memory} combined attention models with reciprocative models to refine a given feature or a set of features. R3Net\cite{deng2018r3net} and RFCN \cite{wang2016rfcn} recurrently fused a saliency map with CNN features or the input image to refine the saliency map. Different from them, we use reciprocative models to iteratively update two kinds of features, \ie, the focal stack feature and the all-focus feature, where the interactions between them are considered to introduce mutual guidance.

\section{Proposed model}
In Figure~\ref{network_overview}, we illustrate the overview of the proposed model. First, we use two encoders to extract features from the all-focus image and the corresponding focal slices, respectively. Then, we input them into the proposed DLG model to propagate contextual features among focal slices, which are further aggregated by the focal stack feature aggregation model. With the proposed reciprocative guidance scheme, focal stack and the all-focus features can be fused with each other several times, hence being improved progressively. Finally, the fused feature is fused with a low-level feature to predict the final saliency map.

\subsection{General GNNs}\label{Section:General GNNs}
GNNs have powerful capability to propagate contexts from neighbouring nodes for graph-structured data. Given a specific GNN model $\mathcal{G}=(\mathcal{V}, \mathcal{E})$, $\mathcal{V} = \{v_{1},v_{2},... ,v_{N}\}$  represents the set of nodes and $e_{i, j} \in \mathcal{E}$ represents the edge from $v_{j}$ to $v_{i}$. Each node $v_i$ has a corresponding node embedding as its inital state $h_{i}^{0}$. We use $\mathcal N_{i}$ to represent the set of neighbouring nodes of $v_i$. GNNs first aggregate contextual information from $\mathcal N_{i}$ to update the state of $v_i$ with a learned message passing function $M$, which has specific formulations in different kinds of GNNs. The general formulation of the message passing process for $v_i$ at step $k$ can be written as:
\begin{equation} \label{eq:message}
m_{i}^{k} = M([h_1^{k-1}, h_2^{k-1}, ...h_j^{k-1}],[e_{i,1},e_{i,2},...e_{i,j},]),
\end{equation}
where each $v_{j} \in \mathcal N_{i}$. After message passing, a state update function $U$ can be learned to update the state of $v_{i}$ based on the aggregated message, which can be defined as:
\begin{equation} \label{eq:update}
h_{i}^{k+1} = U(h_{i}^{k}, m_{i}^{k}).
\end{equation}
Finally, after $K$ steps updating, a readout function can be applied to $h_{i}^{K}$ to get the final output.

\subsection{Feature Encoders} \label{Section:enc}
For the problem of light field SOD, we have an all-focus image $I_a$ and its corresponding focal stack $I_f$ with $N$ focal slices $\{I_{f_{1}}, I_{f_{2}},...I_{f_{N}}\}$, which have different focused regions. Before defining nodes and their embeddings in the graph, we first use encoder networks to extract image features. As shown in Figure~\ref{network_overview}, $I_a$ and $I_f$ are first inputted into two unshared encoders to extract all-focus image features and focal stack features. Similar to previous works \cite{piao2020exploit, zhang2019memory}, we adopt the VGG-19 \cite{simonyan2014very} network without the last pooling layer and fully connected layers as the backbone of our encoder. We obtain high-level features from the last three convolutional stages. Then, we fuse them in an top-down manner \cite{lin2017fpn} and obtain the fused multiscale features $F_{a}\in{\mathbb{R}^{1 \times C \times H \times W}}$ and $F_{f}\in{\mathbb{R}^{N \times C \times H \times W}}$ at the $1/4$ scale, where $F_{f}=\{F_{f_{1}}, F_{f_{2}},...F_{f_{N}}\}$ represents the feature set of $N$ focal slices, $W$, $H,$ and $C$ denote the width, height, and channel number of the feature maps, respectively. 

\subsection{Dual Local Graph}
We use graph models to fuse the focal stack features $F_{f}$ by propagating contexts within the focal slices and also under the guidance from the all-focus feature $F_{a}$. The latter can provide external guidance for the feature update of $F_{f}$.

Directly constructing a densely connected graph among $F_{f}$ and $F_{a}$, which is the case in \cite{wang2019zero,zhang2020adaptive}, requires $(N+1)WH \times (N+1)WH$ edge connections. This scheme is computationally prohibitive for the message passing process when the feature maps have large spatial sizes. The reason for \cite{wang2019zero,zhang2020adaptive} to use densely connected graphs is that the target objects in video object segmentation and co-saliency detection are usually located in different spatial locations. Thus, global context is needed. However, for light field SOD, each all-focus image and its corresponding focal stack images are spatially aligned. Hence, using local context solely is enough. Therefore, in this paper we propose a novel DLG model that only constructs edge connections within local neighbouring nodes for light field SOD. We design two subgraphs, named the focal-focal graph and the focal-all graph, to propagate contextual information from focal slices to focal slices and from the all-focus image to focal slices, respectively.
The whole process can be defined as:
\begin{equation} \label{eq:mlg}
    F_f' = DLG(F_{f}, F_{a}),
\end{equation}
where $F_f'\in{\mathbb{R}^{N \times C \times H \times W}}$ represents the updated feature after context aggregation.

\begin{figure}[!t]
	\graphicspath{{figure/}}
	\centering
	\includegraphics[width=1\linewidth]{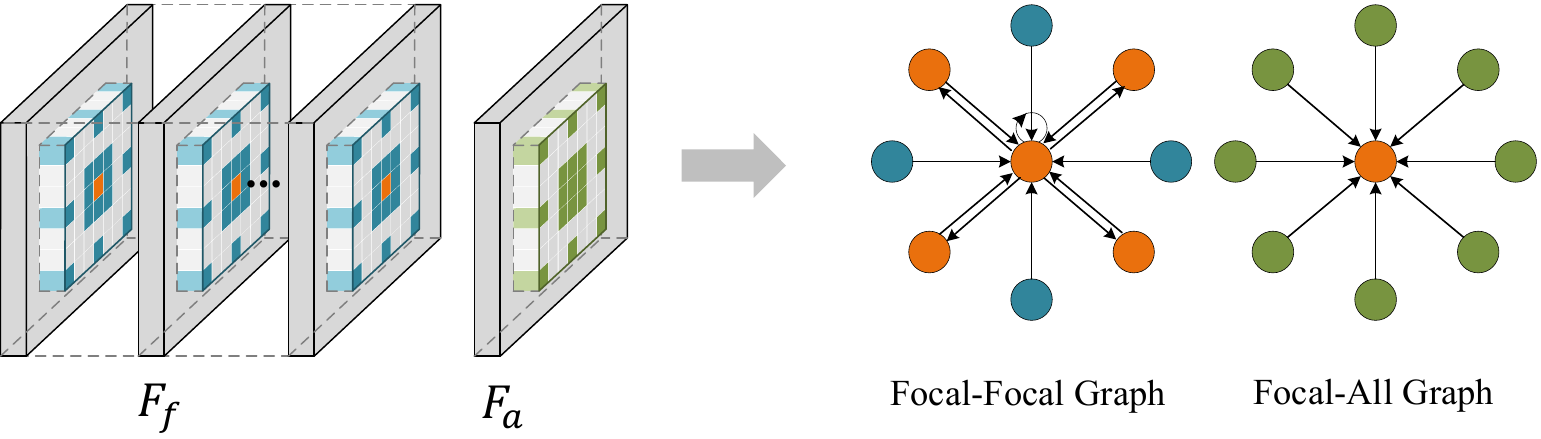}
	\caption{Structure of the Focal-Focal Graph and the Focal-All Graph. For each spatial location $(w,h)$ in $F_f$, we have $N$ target nodes (\textcolor[RGB]{234, 112, 14}{orange}). For their corresponding surrounding area, we have $N \times (k \times k - 1)$ nodes (\textcolor[RGB]{49, 133, 156}{blue}). In $F_a$, we regard the same location $(w,h)$ and its surrounding area as the guidance context and obtain $k \times k$ nodes (\textcolor[RGB]{121, 149, 64}{green}). The connections between one target node and its neighbors in two graphs are shown on the right.} 
	\label{graph}
	\vspace{-0.3cm}
\end{figure}

\vspace{-3mm}
\paragraph{Defination of the surrounding area:}
Before the introduction of the proposed graph network, we first define a surrounding area of a location in a feature map. Given the pixel location $(w,h)$ in a feature map, we have a sampling window with a size of $k \times k$ and dilation $d$, centering at $(w,h)$. Then, we can view all sampled locations except the central one, \ie, the location $(w,h)$ itself, as its surrounding area, as shown as the blue dots in Figure~\ref{graph}. The surrounding area defines the context in a local region and can be used to construct local graph connections.

\vspace{-3mm}
\paragraph{Focal-Focal Graph:}
First, we build a graph $\mathcal{G}^f_{w,h} = (\mathcal{V}^f_{w,h}, \mathcal{E}^f_{w,h})$ for each spatial location $(w,h)$ only in focal features. To ease the presentation, we omit the subscript below. Given the extracted focal stack feature map $F_{f}$, we can view it as having $N$ points from $N$ focal slices with $C$ channels for each spatial location. To be specific, for the location $(w,h)$, we have $N$ target nodes with a $C$-dimensional embedding, which can be defined as $\mathcal{V}_{T}$. In the surrounding area of $(w,h)$, we also have $N \times (k \times k - 1)$ nodes with a $C$-dimensional embedding, where we use $\mathcal{V}_{S}$ to represent the set of these nodes. Here we have $\mathcal{V}_{T} \cup \mathcal{V}_{S} = \mathcal{V}^f$.

After that, we define edges to link these nodes. We follow two rules: \textbf{1)} The nodes in $\mathcal{V}_{T}$ are our modeling targets. Hence they are linked to each other, including themselves. \textbf{2)} The nodes in $\mathcal{V}_{S}$ serves as local contexts for the target nodes. Therefore, they are linked to each node in $\mathcal{V}_{T}$. Except for these edges, there are no other connections in the graph. The two kinds of edges constitute $\mathcal{E}^f$.

Now, we need to define edge embeddings. For simplicity, we use $u$ and $v$ to represent a target node and one of its neighbours, respectively, \ie, $u \in \mathcal{V}_{T}$ and $v \in \mathcal{V}^f$. Their states (features) can be written as $h_u$ and $h_v$. The edge embedding $e_{u, v}^f$ represents the relation from $v$ to $u$. As the two nodes are both from the focal stack feature map, we use inner product to compute the edge embeddings as:
\begin{equation} \label{eq:edge_f}
    e_{u, v}^f = {\theta^f (h_u)}^{\top} \phi^f (h_v),
\end{equation}
where $\theta^f(*)$ and $\phi^f(*)$ are two linear transformation functions
with learnable parameters. They have the same output dimensions and can be implemented by fully connected layers. As a result, 
the computed $e_{u, v}^f $ is a scalar.

\vspace{-3mm}
\paragraph{Focal-All Graph:}
For the target of using the all-focus feature to guide the updating of focal features, we also build a graph $\mathcal{G}^a = (\mathcal{V}^a, \mathcal{E}^a)$ for focal and all-focus image features together at each spatial location $(w,h)$. Again we omit the subscript $(w,h)$ for easy presentation. For location $(w,h)$ in $F_{f}$, we have the same target node set $\mathcal{V}_{T}$. Then, we regard the same location $(w,h)$ in $F_{a}$ and its surrounding area as the guidance context, and obtain a set of $k \times k$ nodes $\mathcal{V}_{S}'$. Here $\mathcal{V}_{T} \cup \mathcal{V}_{S}' = \mathcal{V}^a$.

We connect all nodes in $\mathcal{V}_{S}'$ to each node in $\mathcal{V}_{T}$ to incorporate the guidance context for all target nodes. Here we use $u$ and $q$ to represent a target node and one of its neighbours, respectively, \ie, $u \in \mathcal{V}_{T}$ and $q \in \mathcal{V}_{S}'$. Similarly, their states are denoted by ${h}_{u}$ and $h_{q}$. Since the two nodes are from two different feature spaces, we use a linear transformation to build the edge embedding from $q$ to $u$, which can be defined as:
\begin{equation} \label{eq:edge_a}
    e_{u, q}^a = \psi([{\theta^a (h_u)}, \phi^a (h_q)]),
\end{equation}
where $[,]$ denotes the concatenation operation, $\theta^a(*)$ and $\phi^a(*)$ represent two linear transformation functions. The last linear function $\psi$ projects the input to a scalar.

\vspace{-3mm}
\paragraph{Message passing:}
After getting the embedding for each node and edge, we can define the formulation of the message passing process now. For the target node $u$, we respectively define the message passing in the Focal-Focal Graph and the Focal-All Graph as:

\begin{equation} \label{eq:msg_f}
    m_u^f =\sum_{v \in \mathcal{V}^f} \alpha_{u,v}^f g^{f}(h_v),
\end{equation}

\begin{equation} \label{eq:msg_a}
    m_u^a =\sum_{q \in \mathcal{V}_{S}'} \alpha_{u,q}^a g^{a}(h_q),
\end{equation}
where $g^f(*)$ and $g^a(*)$ are two linear transformation functions in the two Graphs,
and $\alpha_{*}^*$ can be computed by the Softmax normalization:

\begin{equation}
    \alpha_{u,v}^f = \frac{exp(e_{u, v}^f)}{\sum_{j \in \mathcal{V}^f} exp(e_{u, j}^f)},
\end{equation}

\begin{equation}
    \alpha_{u,q}^a = \frac{exp(e_{u, q}^a)}{\sum_{j \in \mathcal{V}_{S}'} exp(e_{u, j}^a)}.
\end{equation}

From \eqref{eq:msg_f} and \eqref{eq:msg_a}, we can aggregate the contexts and guidance from neighbour nodes by considering the relationships inferred in \eqref{eq:edge_f} and \eqref{eq:edge_a}.

\vspace{-3mm}
\paragraph{Node Updating:}
After achieving the messages from neighbours in the two subgraphs, we can update the state of $u$ by:
\begin{equation} 
    h_u' = \varphi^f (m_u^f) + \varphi^a (m_u^a) + h_u
\end{equation}
where $\varphi^f(*)$ and $\varphi^a(*)$ are two linear transformation functions in the Focal-Focal Graph and the Focal-All Graph, respectively, to transform the messages to the original node embedding space.

By adopting the proposed local graph model, the computational complexity of modeling the context propagation in light field images is reduced from $O(((N+1)(HW))^{2}C)$ to $O(NHWC(N+1)k^{2})$. Considering $k^2 \ll HW$, our model shows significant efficiency.

\vspace{-3mm}
\paragraph{Multiscale surrounding area:}
With the introduced surrounding area, the proposed two graph networks can aggregate information in a local region, which can reduce computational costs dramatically. However, only using one sampling window is sensitive to scale variations. Motivated by ASPP  \cite{chen2017rethinking}, we combine multiple sampling windows with different dilation rates to incorporate multiscale and larger contexts, as shown in Figure~\ref{graph}, in which we use two $3 \times 3$ sampling windows with dilation rates of 1 and 3, respectively.

\subsection{Focal Stack Feature Aggregation} \label{Section:Focal Stack Feature Aggregation}
After updating each node embedding $h_u'$ in the DLG model, we can obtain the final output focal stack feature $F_f'$ in \eqref{eq:mlg}. The features of different focal slices have communicated with each other and received guidance from the all-focus feature. Now we can tell useful and useless features among them and aggregate $N$ feature maps into one. First, we use a $1 \times 1$ convolutional layer to reduce the channel number of $F_f'$ from $C$ to $1$. Then, the Softmax normalization function is used along the first dimension to obtain an attention matrix $A \in{\mathbb{R}^{N \times 1 \times H \times W}}$, where the $N$-dimensional attention weights at each location encode the usefulness of each focal slice at this location.
The final aggregated focal stack feature can be obtained by:
\begin{equation} \label{eq:fsfa}
\begin{aligned}
        O &= FSFA(F_f'),\\
          &= \sum_{i=1}^N A^i \odot F_{f_i}',
\end{aligned}          
\end{equation}
where $O\in{\mathbb{R}^{C \times H \times W}}$, $\odot$ is the element-wise multiplication, $A^i$ and $F_{f_i}'$ mean the attention map and the feature map for the $i^{th}$ focal slice, respectively.

\subsection{Reciprocative Guidance}\label{Section:Reciprocative Guidance}
Although the aggregated focal stack feature $O$ can be directly fused with the all-focus feature $F_a$ for predicting the saliency map, we argue that the single-phase fusion scheme can not effectively mine complex interactions and supplements between the two kinds of features, which are crucial for light field SOD. Hence, we propose a reciprocative guidance scheme to make the two kinds of features promote each other for multiple steps.

Here, to avoid confusion, we redefine the outputs from the encoders as $F_{f}^{0}=\{F_{f1}^{0}, F_{f2}^{0},...F_{fN}^{0}\}$ and $F_{a}^{0}$, where the superscripts represent the initial reciprocative step. Then, we define the proposed reciprocative guidance process as:

\begin{equation} 
   F_{f}^{t+1} = DLG(F_{f}^{t}, F_{a}^{t}),
\end{equation}
\begin{equation} 
    O^{t+1} = FSFA(F^{t+1}_{f}),
\end{equation}
\begin{equation} \label{eq:ConvGRU}
    F^{t+1}_{a} = ConvGRU(O^{t+1},  F^{t}_{a}),
\end{equation}
where $t \in [0, T-1]$, $ConvGRU$ is the convolutional gated recurrent unit model \cite{cho2014learning}.

In each step, the all-focus feature $F_{a}^{t}$ is first used to guide the feature fusion of the focal features $F_{f}^{t}$. After the graph model and the feature aggregation, the aggregated focal stack feature $O^{t+1}$ is further used to enhance $F_{a}^{t}$ for saliency detection via ConvGRU. At last, ConvGRU can effectively fuse the two kinds of features, \ie, $O^{t+1}$ and $F_{a}^{t}$, in all reciprocative steps using the memory mechanism. As the reciprocative process goes on, the two features can be improved step by step under the guidance of each other, thus benefiting the final saliency detection. On the other hand, as the reciprocative process goes on, the context propagation in the focal-focal graph can be performed multiple times, hence also enhancing the feature fusion within $F_{f}^{t}$.

\subsection{Saliency Prediction and Loss Function}
Since it has been proved that low-level features can benefit the recovery of object details, we also leverage the low-level all-focus feature to perform saliency map refinement after the reciprocative guidance process. Specifically, we use a skip-connection to incorporate the all-focus feature from the first stage of the encoder VGG network and sum it with the upsampled $F^{T}_{a}$. Then, we perform feature fusion at the 1/2 scale via three $ 3\times 3$ convolutional layers with ReLU activation functions. After that, another $3 \times 3$ convolutional layer with the Sigmoid activation function is used to obtain the final saliency map, as shown in Figure~\ref{network_overview}.

After each reciprocative process, we can obtain an enhanced feature $F_{a}^{t}$. In order to guide our model to gradually enhance the image features, we add a $1 \times 1$ convolutional layer with a sigmoid active function on $F_{a}^{t}$ to predict a saliency map. Then we employ the binary cross-entropy loss to supervise the training of the $t$-th reciprocative step. Finally, the overall loss is the summation of each loss at each step.

\section{Experiments}
\subsection{Datasets}
Our experiments are conducted on three public light field benchmark datasets: LFSD  \cite{li2014saliency}, HFUT \cite{zhang2017saliency}, and DUTLF-FS \cite{wang2019dlsd}. DUTLF-FS is the largest dataset that contains 1462 light field images and is split into 1000 and 462 images for training and testing, respectively. HFUT and LFSD are relatively small, containing only 255 and 100 samples, respectively. Each sample includes an all-focus image, several focal slices, and the corresponding ground-truth saliency map.

\subsection{Evaluation Metrics}
We follow many previous works to adopt the maximum F-measure ($F_{\beta}$)  \cite{achanta2009frequency}, S-measure ($S_{\alpha}$)  \cite{fan2017structure}, the maximum E-measure ($E_{\phi}$)  \cite{fan2018enhanced}, and the Mean Absolute Error (MAE) to evaluate the performance of different models in a comprehensive way.

\begin{table*}[t!]
	\centering
	\scriptsize
	\renewcommand{\arraystretch}{1.2}
	\renewcommand{\tabcolsep}{1.8mm}
	\caption{Benchmarking results. 
		$\uparrow \& \downarrow$ denote larger and smaller is better, respectively. The best scores in each group are indicated in \blu{blue}, and the best scores in all groups are indicated in \red{red}. * means we do not compare with this model due to incompatible training/test split.}
	\begin{tabular}{lrc|cccc|cccc|cccc}
		\hline
		&        &        &\multicolumn{4}{c|}{HFUT \cite{zhang2017saliency}} &\multicolumn{4}{c|}{DUTLF-FS \cite{wang2019dlsd}} &\multicolumn{4}{c}{LFSD \cite{li2014saliency}} \\
		& Methods & Years & $S_{\alpha}\uparrow$ & $F_{\beta}\uparrow$ & $E_{\phi}\uparrow$ & $MAE\downarrow$ & $S_{\alpha}\uparrow$ & $F_{\beta}\uparrow$ & $E_{\phi}\uparrow$ & $MAE\downarrow$ & $S_{\alpha}\uparrow$ & $F_{\beta}\uparrow$ & $E_{\phi}\uparrow$ & $MAE\downarrow$\\  \hline
		
		\multirow{8}{*}{\textit{Light Field}}
		& Ours                        & -    & 0.766 & 0.697 & 0.839 & \red{0.071} & \red{0.928} & \red{0.936} & \red{0.959} & \red{0.031} & \red{0.867} & \red{0.870}  & \red{0.906} & \blu{0.069} \\
		& ERNet \cite{piao2020exploit} & 2020 & \blu{0.778} & \blu{0.722} & \blu{0.841} & 0.082 & 0.899 & 0.908 & 0.949 & 0.039 & 0.832 & 0.850  & 0.886 & 0.082  \\ 
		& MAC  \cite{zhang2020light}   & 2020 & 0.731 & 0.667 & 0.797 & 0.107 & 0.804 & 0.792 & 0.863 & 0.102 & 0.782 & 0.776 & 0.832 & 0.127 \\
		& MoLF \cite{zhang2019memory}  & 2019  & 0.742 & 0.662 & 0.812 & 0.094 & 0.887 & 0.903 & 0.939 & 0.051 & 0.835 & 0.834 & 0.888 & 0.089 \\
		& DLSD \cite{piao2019deep}     & 2019 & 0.711 & 0.624 & 0.784 & 0.111 & * & * & * & * & 0.786 & 0.784 & 0.859 & 0.117 \\ 
		& LFS \cite{li2017saliency}    & 2017 & 0.565 & 0.427 &  0.637 & 0.221 & 0.585 & 0.533 & 0.711 & 0.227 & 0.681 & 0.744 & 0.809 & 0.205 \\
		& WSC \cite{li2015weighted}    & 2015 & 0.613 & 0.508 & 0.695 & 0.154 & 0.657 & 0.621 & 0.789 & 0.149 & 0.700 & 0.743 & 0.787 & 0.151 \\
		& DILF \cite{zhang2015saliency} & 2015 & 0.675 & 0.595 & 0.750 & 0.144 & 0.654 & 0.585 & 0.757 & 0.165 & 0.811 & 0.811 & 0.861 & 0.136 \\\hline
		\multirow{6}{*}{\textit{RGB-D}}
		& BBS \cite{fan2020bbs}  & 2020  & 0.751 & 0.676 & 0.801 & \blu{0.073} & 0.865 & 0.852 & 0.900  & 0.066 & \blu{0.864} & 0.858 & 0.900 & 0.072\\
		& SSF \cite{zhang2020select} &2020 & 0.725 & 0.647 &0.816 & 0.090 & 0.879 & 0.887 & 0.922 & 0.050 & 0.859 & \blu{0.868} & 0.901 & \red{0.067} \\
		& S2MA \cite{liu2020learning} & 2020  & 0.729 & 0.650 & 0.777 & 0.112 & 0.787 & 0.754 & 0.839 & 0.102 & 0.837 & 0.835 & 0.873 & 0.094\\
		& ATSA \cite{zhang2020asymmetric} & 2020  & 0.772 & \blu{0.729} & 0.833 & 0.084 & \blu{0.901} & \blu{0.915} & \blu{0.941} & \blu{0.041} & 0.858 & 0.866 & \blu{0.902} & 0.068 \\
		& JLDCF \cite{fu2020jl} & 2020 & \blu{0.789} & 0.727 & \red{0.844} & 0.075 & 0.877 & 0.878 & 0.925 & 0.058 & 0.862 & 0.867 & \blu{0.902} & 0.070\\
		& UCNet \cite{zhang2020uc} & 2020 & 0.748 & 0.677 & 0.804 & 0.090 & 0.831 & 0.816 & 0.876 & 0.081 & 0.858 & 0.859 & 0.898 &0.072 \\ \hline
		\multirow{6}{*}{\textit{RGB}}
		& LDF \cite{wei2020label} & 2020 & 0.780 & 0.708 & 0.804 & 0.093 & 0.873 & 0.861 & 0.898 & 0.061 & 0.821 & 0.803 & 0.843 & 0.096 \\
		& ITSD \cite{zhou2020interactive} & 2020  & \red{0.805} & \red{0.759} & \blu{0.839} & 0.089 & \blu{0.899} & \blu{0.899} & \blu{0.930} & 0.052 & 0.847 & 0.840 & 0.879 & 0.088\\
		& MINet \cite{pang2020multi} & 2020 & 0.792 & 0.720 & 0.816  & \blu{0.086} & 0.890 & 0.882 & 0.916 &\blu{0.050} & 0.834 & 0.828 & 0.861 & 0.091\\ 
		& EGNet \cite{zhao2019egnet} & 2019 & 0.769 & 0.676 & 0.796 & 0.092 & 0.886 & 0.868 & 0.910 & 0.053 & 0.843 & 0.821 & 0.872 & 0.083  \\
		& PoolNet \cite{liu2019simple} & 2019  & 0.769 & 0.676 & 0.794 & 0.091 & 0.883 & 0.859 & 0.911 & 0.051 & \blu{0.858} & \blu{0.848} & \blu{0.894} & \blu{0.074}\\ 
		& PiCANet \cite{liu2018picanet} & 2018 & 0.783 & 0.715 & 0.816 & 0.107 & 0.876 & 0.865 & 0.907 & 0.072 & 0.832 & 0.834 & 0.866 & 0.103 \\ \hline
		
	\end{tabular}
	\label{tab:Performance_Comparison}
	\vspace{-0.2cm}
\end{table*}

\begin{figure*}[!t]
	\graphicspath{{figure/}}
	\centering
	\includegraphics[width=1\linewidth]{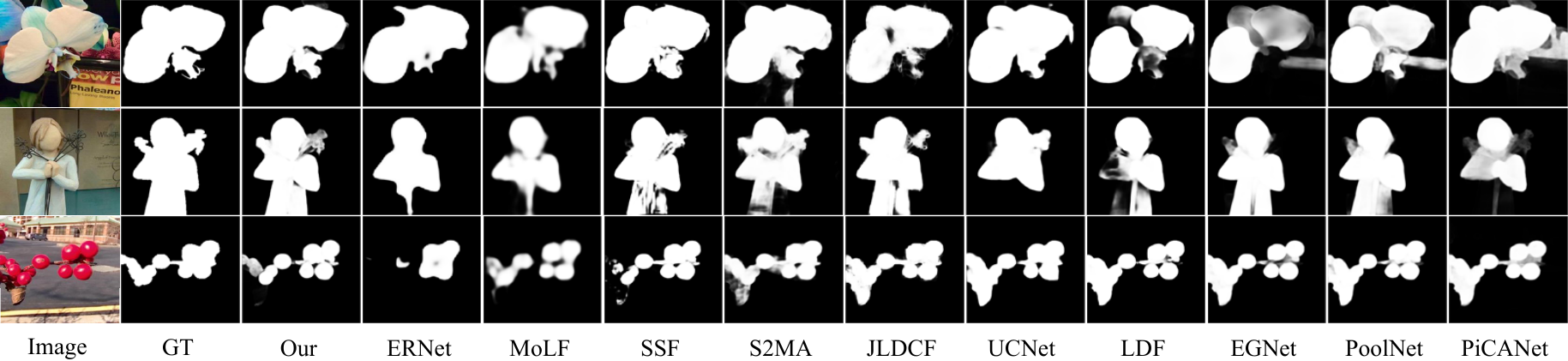}
	\caption{Visual comparison of the saliency maps between our model and state-of-the-art methods.}
	\label{fig:quali_cmp}
	\vspace{-0.4cm}
\end{figure*}

\subsection{Implementation Details}
We design two sampling windows with size $k=3$ and dilation rates $d=1,3$ in DLG, and set the reciprocative step number $T$ as 5 based on experiments. For a fair comparison, we use the same training set with  \cite{piao2020exploit}, which includes the training set of DUTLF-FS and 100 samples selected from HFUT. We also augment the training data with random flipping, cropping, and rotation. We use Adam  \cite{kingma2014adam} as the optimization algorithm and set the learning rate to 1e-4. The minibatch size is set to 1 and our network is trained for 200,000 steps. The learning rate is multiplied by 0.1 at the 150,000 and 180,000 steps, respectively.
In both training and testing, we resize images to $256 \times 256$ for easy implementation. The proposed method is implemented using the Pytorch toolbox \cite{paszke2017automatic} and all experiments are conducted on one RTX 2080Ti GPU. The inference time of our model averaged on all three datasets is only 0.07s per image.
Our code is publicly available at: \url{https://github.com/wangbo-zhao/2021ICCV-DLGLRG}.

\subsection{Comparison with State-of-the-art Methods}
\paragraph{Quantitative Comparison:}
For comprehensive comparisons, we compare our method with 19 state-of-the-art models, including six RGB SOD methods: LDF \cite{wei2020label}, ITSD \cite{zhou2020interactive}, MINet \cite{pang2020multi}, EGNet \cite{zhao2019egnet}, PoolNet \cite{liu2019simple} and PiCANet \cite{liu2018picanet},
six RGB-D SOD models:  BBS \cite{fan2020bbs}, SSF \cite{zhang2020select}, S2MA \cite{liu2020learning}, ATSA \cite{zhang2020asymmetric}, JLDCF \cite{fu2020jl} and UCNet \cite{zhang2020uc},
and seven light field SOD methods:  ERNet \cite{piao2020exploit}, MAC \cite{zhang2020light}, MoLF \cite{zhang2019memory}, DLSD \cite{piao2019deep}, LFS \cite{li2017saliency}, WSC \cite{li2015weighted}, and DILF \cite{zhang2015saliency}.

As shown in Table~\ref{tab:Performance_Comparison}, our method can achieve the best performance on DUTLF-FS and LFSD compared with all RGB, RBG-D, and light field methods. When it comes to the HFUT, our method can surpass other methods in terms of $MAE$, but performs worse in terms of the other three metrics. 
We argue that this is because many images in HFUT have uncommon SOD annotations, such as numbers and texts, which are rarely related to focus information. Thus, our model is not good at handling them.

It is noteworthy that on DUTLF-FS and LFSD, our method significantly outperforms ERNet \cite{piao2020exploit}, MoLF \cite{zhang2019memory} and DLSD \cite{piao2019deep}, which all use ConvLSTM models. This result demonstrates the superiority of our proposed reciprocative scheme. We also note that with only 1100 training samples, our method can achieve better performance on DUTLF-FS and LFSD than most deep RGB-D and RGB SOD methods, which are usually trained on much more images. This indicates that our method can effectively explore the information conveyed in light field data.

\vspace{-0.7cm}
\paragraph{Qualitative Comparison:}
In Figure~\ref{fig:quali_cmp}, we visualize some representative saliency map comparison cases. We can find that, compared with other SOTA methods, our model can not only more accurately localize salient objects, but also more precisely recover object details.

\subsection{Ablation Study}
In this section, we conduct ablation experiments on the largest DUTLF-FS dataset to thoroughly analyze our proposed model.

\begin{table}[]
    \centering
    \footnotesize
    \caption{Quantitative results of using different feature fusion strategies. ``Enc" means our feature encoders in Section~\ref{Section:enc}, ``R" denotes our proposed reciprocative guidance scheme, and ``r" means using the low-level all-focus feature to refine the saliency map. \blu{Blue} indicates the best performance.}
    \begin{tabular}{l|l|cccc}
    \hline
    \multicolumn{2}{l|}{\multirow{2}{*}{Settings}} & \multicolumn{4}{c}{DUTLF-FS} \\
    \multicolumn{2}{l|}{}                      & \multicolumn{1}{l}{$S_{\alpha}\uparrow$} & \multicolumn{1}{l}{$F_{\beta}\uparrow$} & \multicolumn{1}{l}{$E_{\phi}\uparrow$} & \multicolumn{1}{l}{MAE $\downarrow$}   \\ \hline
    
    \multicolumn{2}{l|}{Enc-concat}     & 0.891 & 0.898 & 0.934 & 0.062    \\
    \multicolumn{2}{l|}{Enc-lstm}     & 0.900 & 0.909 & 0.940 & 0.047    \\  \hline
    \multicolumn{2}{l|}{Enc-DLG}   & 0.907 & 0.911 & 0.944 & 0.044 \\ 
    \multicolumn{2}{l|}{Enc-DLG-R}  & 0.923 & 0.932 & 0.957 & 0.035 \\ 
    \multicolumn{2}{l|}{Enc-DLG-R-r}  & \blu{0.928} & \blu{0.936} & \blu{0.959} & \blu{0.031} \\ \hline
    \end{tabular}
    \label{tab:main_ablation}
    \vspace{-0.6cm}
\end{table}

\noindent\textbf{Effectiveness of Different Model Components.}
We first verify the effectiveness of our different model components in Table~\ref{tab:main_ablation}. For fair comparisons, we keep our feature encoders in Section~\ref{Section:enc} unchanged and try different decoder architectures to fuse the focal stack feature $F_f$ and the all-focus feature $F_a$.

We first report the results of two baseline models of fusing $F_f$ and $F_a$ using naive concatenation and LSTM, respectively. For the first one, we follow many previous methods to randomly replicate focal slices in each focal stack to 12 images, then $F_{f}\in{\mathbb{R}^{12 \times C \times H \times W}}$. Next, we concatenate $F_f$ with $F_a$ and use convolution to fuse these 13 feature maps. We denote this strategy as ``Enc-concat". For the second one, we use a ConvLSTM to directly fuse the $N+1$ feature maps in $F_f$ and $F_a$, which is denoted as ``Enc-lstm". We find that using LSTM performs better for fusing the two kinds of features.

Next, we progressively adopt our proposed DLG model, the reciprocative guidance scheme, and the refinement decoder using the low-level all-focus feature. These three models are denoted as ``Enc-DLG", ``Enc-DLG-R", and ``Enc-DLG-R-r", respectively. From Table~\ref{tab:main_ablation}, we can see that the three models can progressively improve the light field SOD performance, finally outperforming the two baseline models by a large margin. Using the DLG model achieves better results than using naive concatenation and LSTM, and also avoids their ill-posed implementation problem. We also try to use a densely connected graph network to fuse the $N+1$ feature maps, but only to obtain the out of memory error. This result proves the efficiency of our DLG model. Furthermore, we find that the reciprocative guidance scheme brings the largest model improvement, clearly demonstrating its powerful capacity. We believe this strategy can also benefit future light field SOD research a lot.

\begin{figure}[!t]
    \graphicspath{{figure/}}
    \centering
    \includegraphics[width=1\linewidth]{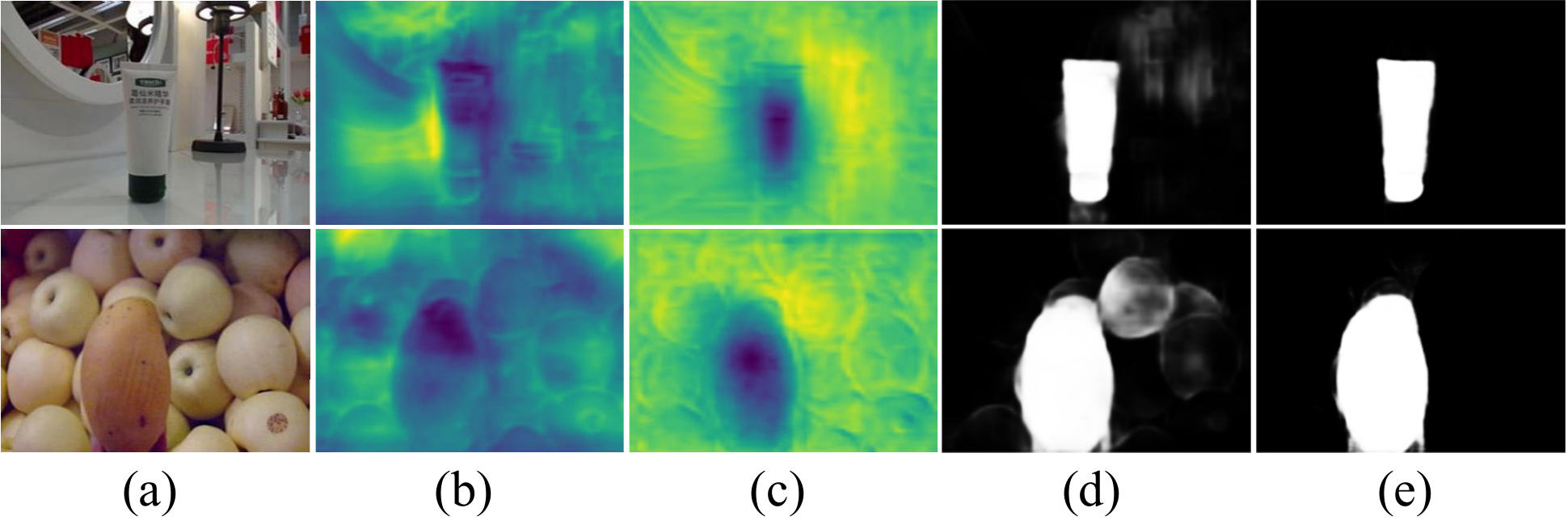}
    \caption{Comparison of models with and without using the proposed DLG model in terms of both feature maps and saliency maps. (a): Images. (b): Feature maps w/o using DLG. (c): Feature maps w using DLG. (d): Saliency maps w/o using DLG. (e): Saliency maps w using DLG.} 
    \label{ablation:DLG}
    \vspace{-0.4cm}
\end{figure}

We also show the comparison of the feature maps and saliency maps with and without using the DLG model in Figure~\ref{ablation:DLG}. We can see that by using DLG, the feature maps can filter out distractions in backgrounds and focus more on the salient objects, hence resulting in better SOD results.

\noindent\textbf{DLG Settings.}
Since we build multiscale local neighbours in DLG to introduce larger contexts with acceptable computational costs, we also explore different multiscale settings in DLG in Table~\ref{tab:dlg_ablation}. Specifically, we test different settings of the sampling window size $k$ and the dilation rates $d$ in the ``Enc-DLG-R" model. We start from the naive setting with $1\times 1$ sampling window. From Table~\ref{tab:dlg_ablation}, we can find that, when we use more and larger windows, the performance of our model can be gradually improved. However, the performance is saturated when using two $3\times 3$ sampling windows with dilation rates of 1 and 3. Further using one more window with $d=5$ only brings little improvement. Considering the computational costs, we choose $k=3$ and $d=1,3$ as our final setting.

To verify the effectiveness of simultaneously using the focal-focal graph $\mathcal{G}^f$ and the focal-all graph $\mathcal{G}^a$, we try to use them separately and report the results in the last two rows in Table~\ref{tab:dlg_ablation}. We find that using them separately will degrade the model performance, hence verifying the necessity of our proposed dual graph scheme.

\noindent\textbf{Reciprocative Steps.}
We conduct experiments to choose the optimal reciprocative step number $T$ in Table~\ref{tab:reci_ablation}. Note that, when $T=1$, the model downgrades to the ``Enc-DLG" model. We find that, as we increase $T$ from 1 to 5, the performance can be improved progressively. When $T>5$, we observe that the performance has been saturated and the model will exceed the GPU memory. Hence, we take $T=5$ as the final setting for our reciprocative guidance scheme.

\begin{figure}[!t]
    \graphicspath{{figure/}}
    \centering
    \includegraphics[width=1\linewidth]{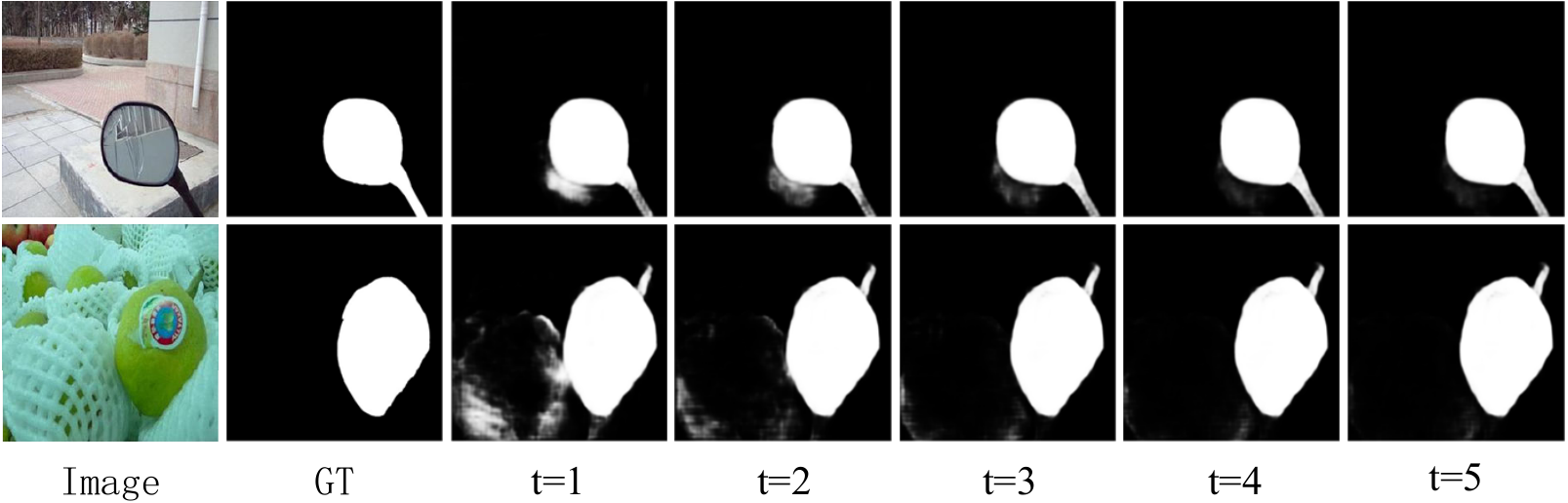}
    \caption{Visualization of the saliency maps at different reciprocative steps.} 
    \label{ablation:reciprocative}
    \vspace{-0.6cm}
\end{figure}

We also visualize two representative samples to show the improvements of the saliency maps obtained at different reciprocative steps in Figure~\ref{ablation:reciprocative}. We can find that, along with the reciprocative guidance learning, false-positive highlights can be gradually suppressed and the SOD results can be steadily improved.

\begin{table}[]
    \centering
    \footnotesize
    \caption{Comparison of using different DLG Settings.}
    \begin{tabular}{l|l|l|l|cccc}
    \hline
    \multicolumn{4}{c|}{Settings} & \multicolumn{4}{c}{DUTLF-FS} \\ \hline
    $k$ &  $d$  & $\mathcal{G}^f$  & $\mathcal{G}^a$  & \multicolumn{1}{l}{$S_{\alpha}\uparrow$} & \multicolumn{1}{l}{$F_{\beta}\uparrow$} & \multicolumn{1}{l}{$E_{\phi}\uparrow$} & \multicolumn{1}{l}{MAE $\downarrow$}   \\ \hline
    
    1  &  1      & \ding{51} &  \ding{51}     & 0.914 & 0.919 & 0.946 & 0.042 \\
    3  &  1      & \ding{51} &  \ding{51}     & 0.915 & 0.919 & 0.950 & 0.038  \\
    3  &  1,3    & \ding{51} &  \ding{51}     & 0.923 & 0.932 & 0.957 & \blu{0.035} \\
    3  &  1,3,5  & \ding{51} &  \ding{51}     & \blu{0.924} & \blu{0.933} & \blu{0.960} & \blu{0.035} \\ \hline
    3  &  1,3    & \ding{51} &  \ding{55}     & 0.917 & 0.929 & 0.953 & 0.038 \\
    3  &  1,3    & \ding{55} &  \ding{51}     & 0.919 & 0.927 & 0.952 & 0.037 \\
    \hline
    \end{tabular}
    \label{tab:dlg_ablation}
    \vspace{-0.2cm}
\end{table}

\begin{table}[]
    \centering
    \footnotesize
    \caption{Comparison of using different reciprocative step numbers.}
    \begin{tabular}{l|cccc}
    \hline
    \multirow{2}{*}{$T$} & \multicolumn{4}{c}{DUTLF-FS} \\
                 & \multicolumn{1}{l}{$S_{\alpha}\uparrow$} & \multicolumn{1}{l}{$F_{\beta}\uparrow$} & \multicolumn{1}{l}{$E_{\phi}\uparrow$} & \multicolumn{1}{l}{MAE $\downarrow$}   \\ \hline
    
    1     & 0.907 & 0.911 & 0.944 & 0.044 \\
    3     & 0.917 & 0.924 & 0.952 & 0.040  \\
    5     & \blu{0.923} & \blu{0.932} & \blu{0.957} & \blu{0.035} \\
    \hline
    \end{tabular}
    \label{tab:reci_ablation}
    \vspace{-0.2cm}
\end{table}

\section{Conclusion}

In this paper, we propose a novel dual local graph neural network and a reciprocative guidance architecture for light field SOD. Our DLG model efficiently aggregates contexts in focal stack images under the guidance of the all-focus image. The reciprocative guidance scheme introduces iterative guidance between the two kinds of features, making them promote each other at multiple steps. Experimental results show that our method achieves superior performance over state-of-the-art RGB, RGB-D, and light field based SOD methods on most datasets. 

\vspace{-4mm}
\paragraph{Acknowledgments:}
This work was supported in part by the National Key R\&D Program of China under Grant 2020AAA0105701, the National Science Foundation of China under Grant 62027813, 62036005, U20B2065, U20B2068.

{\small
\bibliographystyle{ieee_fullname}
\bibliography{egbib}
}

\end{document}